# A primer on synthetic health data

Jennifer A Bartell*[1], Sander Boisen Valentin[2], Anders Krogh[1,3], Henning Langberg[4], and Martin Bøgsted[2,5]

[1]Center for Health Data Science, University of Copenhagen, Copenhagen, Denmark
[2]Department of Clinical Medicine, Aalborg University, Aalborg, Denmark
[3]Department of Computer Science, University of Copenhagen, Copenhagen, Denmark
[4]Department of Public Health, University of Copenhagen, Copenhagen, Denmark
[5]Center for Clinical Data Science, Aalborg University Hospital and Aalborg University, Aalborg, Denmark

*corresponding author

**Abstract**
Recent advances in deep generative models have greatly expanded the potential to create realistic synthetic health datasets. These synthetic datasets aim to preserve the characteristics, patterns, and overall scientific conclusions derived from sensitive health datasets without disclosing patient identity or sensitive information. Thus, synthetic data can facilitate safe data sharing that supports a range of initiatives including the development of new predictive models, advanced health IT platforms, and general project ideation and hypothesis development. However, many questions and challenges remain, including how to consistently evaluate a synthetic dataset's similarity and predictive utility in comparison to the original real dataset and risk to privacy when shared. Additional regulatory and governance issues have not been widely addressed. In this primer, we map the state of synthetic health data, including the latest generation and evaluation methods and tools, existing examples of deployment, the regulatory and ethical landscape, access and governance options, and opportunities for further development.

**Introduction**

Synthetic data has long been a standby of health IT infrastructure projects and educators who need a safe-to-share stand-in for real, person-sensitive health data. Theoretical models built from first principles and known relationships are often used to simulate data that technically would qualify as synthetic, but we are interested in a more modern definition of synthetic data that uses data-driven machine learning approaches to capture the patterns of complex, noisy real-world health data. The latter involves fitting statistical or algorithmic models to real-world datasets, which can then be drawn from to generate synthetic datasets that align with the characteristics of the real-world dataset[1]. If generated properly, these modern synthetic datasets can offer significant advantages over datasets which have been anonymized via removal, scrambling, or binning of sensitive data. Such synthetic datasets aim to support scientific research by enabling project design, analysis method and tool prototyping, and validation of published findings by researchers without access to the underlying sensitive dataset. However, to support high-quality research outcomes, high-quality synthetic datasets should be produced. This has historically been quite a challenge with respect to the large, irregular, and incomplete datasets often encountered within the healthcare sector.

With the advent of generative deep learning, the potential of high-quality synthetic data generation has rapidly expanded. These data-driven methods aim to maintain the same joint distribution in the synthetic data as in the original data and can replicate complex multidimensional patterns and nonlinear relationships with greater success and ease than prior methods. Producing high quality synthetic data is the delicate art of generating samples that lie close to the observed data without getting so close that privacy is endangered. Methods to assess and/or mitigate risk to privacy have thus sprung up alongside the growth of advanced synthesis techniques. These developments are opening the door to much more sophisticated uses of synthetic health data as shown in Table 1. A primary interest is in expanded sharing of health data in a safe manner for research and development.

**Table 1.** Applications enabled by the increase in quality of synthetic health data as data-driven modeling approaches have advanced.

| Application | Intended users | Main hurdle to address |
| --- | --- | --- |
| Hypothesis generation before full data access | Qualified researchers | Need to govern synthetic dataset sharing depending on privacy concerns |
| Prototyping of models and methods before full data access | Qualified researchers | Need to govern synthetic dataset sharing depending on privacy concerns |
| Clinical trials: from data-driven design to synthetic control arms | Qualified researchers | Need to effectively account for any biases in synthetic datasets |
| IT platform development challenged by flaws of real-world data | Health sector & biotech IT / software engineers | Size of needed synthetic datasets might be computational expensive |

| Reproducible study results without sharing private datasets | From qualified researchers to the general public | Requires high quality synthetic data production and validation by originating research team |
| --- | --- | --- |
| Extension of limited datasets (i.e. in cases of rare disease) and bias correction (hybrid synthetic-real datasets) | Qualified researchers | Need to effectively account for any biases in synthetic datasets |
| Probing analysis flaws and tool limitations (designed/directed synthetic data) | Researchers and data science trainers/trainees | Limited hurdles exist given data manipulation focus (lower privacy risk) |
| Education and training in a real-world context without sharing sensitive data with trainers and trainees | Data science trainers/trainees | Need to govern synthetic dataset sharing depending on privacy concerns |

**Factors in data access under the GDPR**

It is generally agreed that increasingly complex data protection legislation such as the EU's General Data Protection Regulation (GDPR) has seriously slowed and constrained the assembly, analysis, and sharing of large-scale health datasets. Yet these are the datasets on which many modern machine learning methods capitalize. Synthetic health data may facilitate compliance with such regulations as generated datasets should erase any one-to-one link between real and synthetic patients (Figure 1). However, with the advent of deep learning approaches that are theoretically capable of learning and reproducing real data in the generated datasets (amongst other concerns), this picture gets complicated.

> **Information Box 1.** The GDPR governs all personal data collection, processing, and sharing within the EU, generally requiring informed user consent throughout these steps and carving out additional safeguards for particularly sensitive data such as personal health data (described in Article 9). Personal health data access without individual consent is allowed via a few exceptions including for scientific research, as long as the research subject's fundamental rights and interests are safeguarded the policy goal of creating a harmonised regulatory framework for health research, the right to data protection is respected, and degree of data processing is proportionate to research aims [Mitchell and Hill, 2023]. The GDPR provides additional context for determining whether natural persons are identifiable and thus data has failed to be anonymized sufficiently for non-applicability of the GDPR (Recital 26) or will be impacted by information disclosure via loss of statistical confidentiality (Recital 162) during sensitive data processing. Specific standards for sensitive health data safeguarding that meet the GDPR research clause's stipulations are interpreted by national authorities via additional legislation [Vlahou et al. 2021].

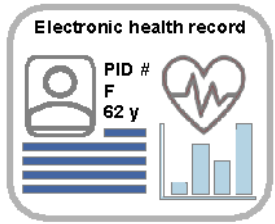

**Dataset: patient attributes of a genetic disease cohort**
Patient ID: national citizen number used by public sector
Health: asymptomatic vs **symptomatic**
Sex: F or M
Mutation: none, variant X in Gene Z, variant Y in Gene Z
Age: 60 – 80 years

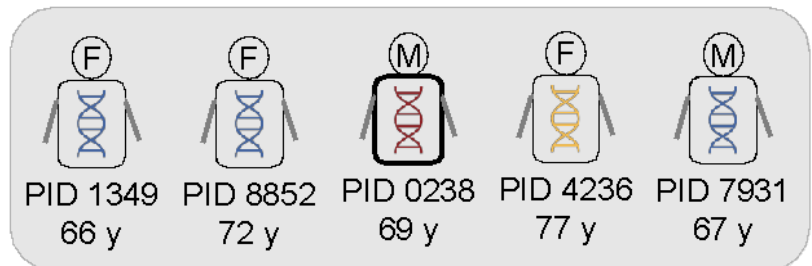

**Characteristics**
Sex: 60% F
Health: 20% symptomatic
Mutation: 20% with variant X, 20% with variant Y
Sensitive PID: yes
Age: 70.2 ± 4.4

Disease predictor: **variant Y**

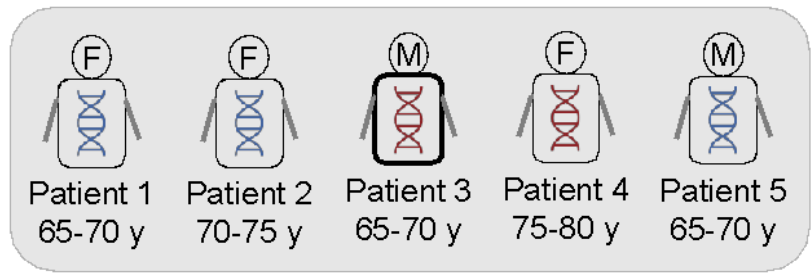

**Characteristics**
Sex: 60% F
Health: 20% symptomatic
Mutation: 40% with mutation in Gene Z
Sensitive PID: no
Age: 60% in [65-70]

Disease predictor: **???**

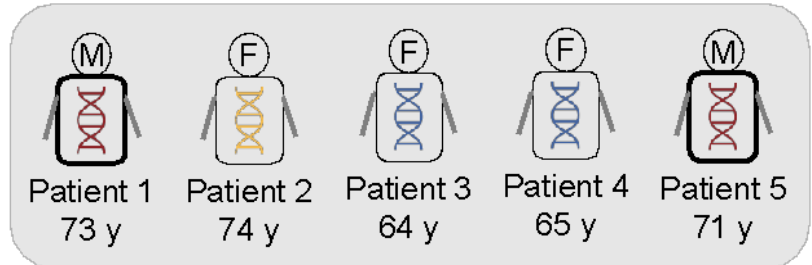

**Characteristics**
Sex: 60% F
Health: 40% symptomatic
Mutation: 20% with variant X, 40% with variant Y
Sensitive PID: no
Age: 69.4 ± 4.6

Disease predictor: **variant Y**

**Figure 1.** A comparison of a mock dataset describing the characteristics of a patient cohort with a disease that has a genetic driver. Real patients are described by a personal identification number (PID), sex (M or F), age in years, health (symptomatic patients have a bolded outline), and mutation status (DNA symbol colored blue for no variant, yellow for variant X, and red for variant Y). Potential research conclusions that could be derived from each dataset are represented by the 'Disease predictor' category alongside the extracted Characteristics summarized on the left. While signal is lost alongside the sensitive data removed during patient anonymization, the careful generation of new "synthetic" patients better preserves patient characteristics as well as study conclusions.

Here, we use the GDPR as the most stringent standard for international regulation of sensitive personal data[§] that includes regulations for health data research in Article 9[2,3]. While non-EEA nations may not yet have implemented data protection at the stringency level of the GDPR, a raft of new data protection regulations have been passed or are under development, with notable efforts in Brazil, Canada, China, and at various state levels in the US[2,4]. Information Box 1 provides a summary of how the GDPR relates to research with sensitive personal health data. While anonymized datasets are outside the GDPR's purview, pseudonymized datasets are still considered personally sensitive and under the jurisdiction of the GDPR because patients can theoretically be re-identified or sensitive information about them can be disclosed.

*Anonymization*
The GDPR considers a dataset anonymous and excluded from regulation when a natural person cannot be directly or indirectly identified by means reasonably likely to be used with respect to cost, time, and current and developing technology[5]. Anonymization is typically performed by grouping data to ensure non-uniqueness of identifying or quasi-identifying values across at least $k$ individuals ($k$-anonymity[6]), sufficient variation of certain variables (L-diversity[7]) and similarity of probability distributions (t-closeness[8]). Amongst many developed approaches[9], k-anonymity remains the de-facto minimum acceptable standard of anonymisation according to the European Medicines Agency[10]. However, it has been shown to decrease the utility of data[11,12], is vulnerable to several forms of attack[7,8], and is actually computationally challenging to perform given the variable combinatorics[13]. A major struggle in this approach is defining what variable sets represent a current or future re-identification risk. In general, performing verifiable, future-proof anonymization that retains sufficient data utility has grown increasingly difficult.

*Pseudonymization*
In the health sector, pseudonymized health data is commonly used for clinical research under regulatory safeguards implemented at local, national, and international levels (i.e. the GDPR within the EU). Sensitive datasets are coded with a stand-in patient identifier while names, citizen identification numbers, and other identifying information is removed to protect privacy. However, the pseudonymization key linking identifier to real patient is securely retained rather than being erased, which allows further data integration and follow-up studies as necessary. Even if the pseudonymization key were erased, it might be possible to re-identify patients without further privacy enhancing steps such as grouping identifying data (e.g., age, location) into bins. Thus, pseudonymized data must be handled under much more rigorous safeguards than anonymized data, with substantial restrictions on who is allowed to access it and how and where it is stored.

*Synthetic data with respect to sensitive data protection*
Appropriate regulation of synthetic health data is currently an open question because it originates from 'personal data' governed by the GDPR, and so any downstream output will also likely be considered 'personal data' unless paired with effective anonymization that assures negligible danger of privacy attacks or information disclosure[5]. To perform this anonymization step with confidence requires agreement on synthetization standards and effective privacy risk assessment by practitioners, data regulators and policymakers[5,14] – an uphill climb given the rapid pace of technological advancements and fear of accidents, attacks, privacy scandals, and fines. The field has started out by deriving quantitative estimates of risk to privacy from anonymization standards, which we discuss below. Thought leaders in the field have also outlined key considerations for regulatory bodies developing synthetic data policy[15]. To our knowledge, under GDPR jurisdiction, only the United Kingdom[16] has provided public guidance around synthetic data generation and deployment from official regulatory bodies.

Regardless of whether the final synthetic dataset is classed as fully anonymous, the generation of synthetic health datasets is an act of data processing that must fall under an approved category according to the GDPR (or similar regulations). While other GDPR clauses may support synthetic dataset generation, it is easy to argue that synthetic health dataset generation falls under the guidance for scientific research if: 1) methods and use cases are being developed as an active methodological research project with results disseminated to the research community as usual and 2) data access, processing, and research aims regarding synthetic data generation have been approved by relevant data authorities for research purposes.

Ideally, sharing synthetic health data should be safer than sharing pseudonymized or even anonymized datasets, but that does not mean it is risk-free from a privacy perspective. In addition to use of quantitative evaluation metrics that illustrate minimal risk to privacy, any form of release of a synthetic health dataset should be pre-empted by a cost-benefit analysis that addresses the following practical and ethical issues:
1) The original dataset may contain 'extra sensitive' data or have too many structural flaws to be worth the generation effort or risk of releasing the produced synthetic dataset.
2) Future risks related to adversarial attacks on synthetic datasets will likely expand and requires thorough consideration of safeguards and use of up-to-date evaluation techniques.
3) Stakeholders (data authorities) need to consider costs of data support (synthetic and otherwise) when releasing data – beyond potential privacy risks, this includes how publicly funded resources might be supported and how patients might benefit (directly or indirectly) when their data is used.

Ultimately, a synthetic dataset project needs to produce enough positive outcomes and societal benefits to be worth the (actively minimized) risks to privacy – this is the core issue to address in light of the GDPR and general ethical standards.

**Generating useful synthetic datasets**

Empirical methods for generating synthetic datasets generally aim at modeling the statistical characteristics of a population and drawing samples from estimated models to create synthetic patients. We find it useful to group prominent data-driven methods based on their ability to replicate both marginal distributions (i.e., characteristics and distributions within a single variable) and multivariate dependence structures of varying complexity (i.e., joint distributions and correlations across variables). Methods classification is challenging due to compounding improvements and cross-method integrations, so we propose four broad categories:
    1) sequential conditional modeling – a system of sequential regressions or classifications (including sequential trees and imputation methods such as multiple

imputation by chained equations) which allows estimation of the marginal distributions but does not guarantee estimation of the full joint distribution and is influenced by the modeling order of variables[17–19]

2) copulas – a (usually) parametric approach to defining the cumulative distribution function which allows estimation of the joint distribution given continuous marginal distributions but is mathematically formidable and dependent on continuous input data and choice of underlying copula model[20,21]

3) Bayesian networks – probabilistic graphical models of dependencies which can in principle estimate the full joint distribution, infer marginal distributions and offer insights into causal relationships, but struggle with complex and nonlinear dependencies and mixed data types[22]

4) generative deep learning – methods such as generative adversarial networks[23] and variational autoencoders[24] that are capable of implicitly learning the full joint distribution as well as marginal distributions and can capture complex and nonlinear dependencies given sufficiently large datasets and training time.

A more detailed review of specific methods within each group is beyond the scope of this paper, and we refer readers to recent reviews[22,25,26] for a more thorough discussion. We also must mention a parallel concept, differential privacy, which can be integrated in data-synthesis algorithms[27,28]. Differential privacy is a formal mathematical guarantee that there is no increased risk of being harmed from being included in a dataset[29]. An attack would draw the same or negligibly different conclusions about a person's sensitive information based on the dataset, whether the person participates or not. An important weakness with this approach is that outliers, which may be critical to derived outcomes from the real dataset, are likely to be discarded in the synthetic data which will have an effect on its usefulness.

In practice, most real-world datasets will each have their own unique challenges in the synthesis process. Currently the ability to fully automate the production of high-quality synthetic datasets is limited by missing values and overly sparse datasets, collinearity and imbalance, too many variables versus number of patients, and other idiosyncrasies and custom solutions are often required before or during synthesis. However, the many generative deep learning methods in active development and expansion have already begun to address these issues and are a tremendous new resource for creating high quality synthetic health data. The true challenge is in effectively evaluating the produced datasets such that their strengths and weaknesses are clear to downstream users.

**Evaluating synthetic datasets**

Before deploying a synthetic dataset, quantitative and qualitative evaluation metrics must be applied to determine the success of dataset synthesis and suitability for the targeted use cases for the synthetic data. Synthetic data should be evaluated for three overarching qualities:
1) Fidelity - the closeness of the synthetic dataset in scope, structure, and multivariate relationships to the original dataset which at minimum is measured by statistical comparisons of univariate and multivariate distributions
2) Privacy - the degree of potential disclosure of patient identity and sensitive information that could be derived from the synthetic dataset which is commonly evaluated by metrics of potential for re-identification of a person or inference of their attributes or membership in the training data
3) Utility – the ability to replicate predictions and analysis outcomes from the original sensitive dataset with the synthetic dataset which is commonly evaluated in line with prior analyses of the original sensitive dataset

Fidelity and utility are partially overlapping qualities, but separate evaluations should not be neglected and tailored to intended uses of the synthetic dataset – a high utility score for a single particular outcome would be useful for teaching purposes though the synthetic dataset's fidelity remains low and privacy remains high. A trade-off where increased synthetic dataset fidelity results in decreased privacy is generally acknowledged, with new methods like the synthetic data modeling aiming to reduce privacy risk while maintaining fidelity. The generation methods discussed in the prior section have been ordered such that they roughly align along the privacy-fidelity trade-off, whereas model complexity grows (from sequential synthesis to deep learning methods), fidelity of the synthetic data increases and privacy preservation decreases. The appropriate balance of these qualities is dependent on the intended use of a synthetic dataset, and a quantitative evaluation of tradeoffs between these qualities for each synthetic dataset is integral to realistically assessing the risk of sharing it.

*The most promising evaluation metrics to date*
There is currently a large research focus on quantitative metrics of privacy, fidelity, and of synthetic health datasets. We recommend the reviews and comparative analyses of commonly used metrics by Goncalves et al.[22] and Hernandez et al.[30] It is worth highlighting new methods developed by the van der Schaar research group which allow rigorous sample-level estimations of precision, recall, and authenticity of synthesized data that provide much greater information about generative model performance than prior metrics[31].

Ultimately, privacy preservation is a question of risk assessment under reasonable forms of attack rather than guarantee against identity or attribute disclosure. Assuming that no information of the model's construction is publicly disclosed, fully synthetic data should theoretically be anonymous and safe to share. However, there are techniques and attacks to infer a person's probable membership within a dataset with varying prior knowledge[32–35]. There are multiple approaches like attribute matching [36,37], shadow model approaches [32,38] and Bayesian statistical methods[39,40]. A framework specifically rooted in the GDPR has been proposed by El Emam et. al. for evaluating the risk of learning "something new" from synthetic datasets given a correct identification of an individual for discrete variables and discretized continuous variables, relying on complete prior knowledge of quasi-identifiers to estimate the risk of accurate identification[37]. These methods all have their strengths and weaknesses, but no metrics have been broadly accepted as compliant with the wording of the GDPR for release of synthetic data.

**Toolsets for developing synthetic health datasets**

The rapidly growing list of methods for generating and evaluating synthetic datasets is overwhelming for newcomers to the field. However, user-friendly software libraries of synthesis and evaluation tools are beginning to emerge. There are quite a number of options for generating synthetic data[41], but few toolsets address the specific demands of synthetic health data. For example, realistic ranges for physiologic measures from lab tests are broadly available and means to effectively constrain synthesized values to these ranges is therefore a priority in synthetic health data. Furthermore, substantial prior knowledge is often available about trends and relationships within health data, such as hard constraints on gender-specific diseases as well as a large body of knowledge on likely correlations between different phenotypes[42]. Means to incorporate this prior knowledge into the generation of synthetic samples is therefore also a focal point of method development. Below, we describe two comprehensive tool libraries that address these health data-specific needs in synthetic data generation and evaluation.

The Synthetic Data Vault (SDV)[43] is a flexible and well-documented platform for exploring data synthesis that offers auxiliary features that are especially relevant for addressing health datasets, such as means of constraining the ranges of specific variables and filtering out unwanted values. This Python module provides at least one method suited for most data synthesis scenarios (ranging from copulas to CTGAN[44]), and has started to incorporate evaluation metrics (SDMetrics) and a benchmarking platform (SDGym). While offering a genuine "plug-and-play" solution, evaluation metrics deployed thus far are limited especially with respect to privacy, and the SDV suite accessible on GitHub is only available for non-production-scale use (via a Business Source License) as the developers are also marketing a commercial SDV Enterprise product.

To date, the most comprehensive open source toolset for generating synthetic data is a Python library called Synthcity[41] which is under active development and expansion on GitHub. It contains a full modular suite of the latest generation methods and evaluation metrics addressing fidelity, utility, and privacy[41], and includes an implementation of the sample-level approach to evaluating precision, recall, and authenticity of synthetic datasets mentioned earlier[31]. The developers are prominent academic specialists in synthesis of sensitive health data, and the package has been well equipped for this task though it is designed to apply to diverse synthetic data scenarios. It handles challenging input data such as composite datasets (made from mixed static and temporal data). It has 19 synthesis method plug-ins included to date, which are an array of generative deep learning approaches for data generation. Furthermore, it has extensive tools for evaluating machine learning fairness and bias as well as privacy.

**Tabular synthetic health datasets released in the European research space**

Despite the many resources we have discussed for generating and evaluating synthetic data, very few of these datasets have been generated from sensitive real-world health data and shared with the research community. The many technical papers developing methods and metrics rely on standard open-source benchmark datasets, and while these methods are at times applied to private sensitive datasets[45], the produced synthetic datasets are not released alongside the study. In Table 2, we present the tabular synthetic health datasets which have been produced for datasets under the jurisdiction of the GDPR (or GDPR-adjacent legislation in the UK) and been made accessible to the research community at minimum. None of the five datasets were generated using deep learning approaches and four of the datasets were generated in the United Kingdom, which has its own data protection legislation modeled on the GDPR. Furthermore, any datasets with completely unrestricted, freely downloadable access also explicitly meet the GDPR's requirements for anonymized data.

**Table 2.** Tabular synthetic health datasets produced within Europe under the jurisdiction of stringent data protection law.

| Dataset | | | |
|---|---|---|---|
| **Data scope** | **Generation method** | **Privacy preservation method** | **Access & use cases** |
| *Simulacrum (v2.1.0).* Frayling. 2018.[46] https://simulacrum.healthdatainsight.org.uk/wp/wp-content/uploads/2018/11/Methodology-Overview-Nov18.pdf | | | |
| *Synthetic patients:* 1,871,605<br><br>*Data overview*: 122 variables across 9 linked datasets for patient, tumour, treatment regimen, cycle, drug details, and outcomes<br><br>*Data types*: categorical, discrete, and continuous | **Bayesian network:** Directed acyclic graph (DAG) are built from strong pair-wise correlations (via chi-sq. statistic) found in record stratas. These are converted into a Bayesian network where nodes (characteristics) are conditioned on parent nodes to produce a probability distribution from which synthetic values are randomly sampled. | **K-anonymity:** Preprocessing steps enhance privacy by first building DAGs within strata defined by tumour cancer site and/or treatment type and made up of at least 1000 patient records. K-anonymity of minimum k = 50 records per node (characteristic) value is enforced by clustering records of less populated node values until k = 50 is reached. | Open access via form request as data meets anonymization standards |
| *Synthea Covid-19.* Walonoski et al. 2020.[47] | | | |
| *Synthetic patients:* 124,150<br><br>*Data overview*: 77 variables across 16 linked data tables covering COVID19 infection, symptoms, severity, morbidity outcomes, and consumption of PPE and other medical devices (e.g. ventilators)<br><br>*Data types*: categorical, | **Probability-based logic model/state machine workflow**: State transition machines are used to configure temporal models to replicate patient treatment histories based on public demographic information. Model logic and stepwise rules are informed by publicly available patient statistics and clinical care maps rather than learned correlations from sensitive data. | **Derived from population statistics:** Data are generated from aggregated summary statistics from publications rather than using person-specific sensitive information. | Open access from SyntheticMass<br><br>Intended for health IT projects and estimates of medical equipment needs; limited utility for research |
| discrete, and continuous values | | | |
| *CRPD Covid-19 (v4).* Khorchani et al. 2022.[48] | | | |
| *Synthetic patients:* 4,173,000<br><br>*Data overview*: 49 variables across 6 linked data tables, covering COVID19 infection, symptoms, medication, co-morbidities, and patient background<br><br>*Data types*: categorical outside of dates and patient age | **Random draws from population statistics with enforced correlations**: Public patient statistics were used to define distributions for value draws. A public anonymized dataset was the source of constraining rules for single-parameter distributions and their relative correlations. These rules were enforced using a linear decomposition method (related to Cholesky decomposition in a subset of variables). | **Derived from public datasets:** Synthetic data created via public data is not necessarily privacy preserving, but the synthesizers' culpability in information disclosure is likely limited under the GDPR. They call their approach SASC – a simple approach to synthetic cohorts. | 200 GBP for research access, 1100 GBP for teaching access |
| *CRPD Cardio-vascular dataset (v1).* Wang et al. 2021.[42] | | | |
| *Synthetic patients:* 499,344<br><br>*Data overview*: 23 variables across 5 linked data tables covering cardiovascular symptoms, co-morbidities, smoking status, and patient background<br><br>*Data types*: continuous and categorical | **Bayesian network:** A directed acyclic graph (DAG) structure is used to build a Bayesian network made up of nodes with local conditional distributions from which synthetic values are randomly sampled. Then, key biological relationships, univariate and bivariate distances, and privacy concerns are preserved as 'ground truths' in semi-automated fashion. Sample synthetic datasets were compared in batches to the same size sample of sensitive data, and measures of 'ground truth' were compared for similarity to | **Enforced via ad-hoc filtering and outlier analysis:** Sensitive variables were pre-defined based on official guidance and standards from the UK Information Commissioner's office (ICO), though it is not completely clear how these were used. 'Identical data instances' were a privacy concern if these instances were also 'rare' according to ground truth. A density-based clustering algorithm (DBSCAN) was used to identify these outliers. If no identical outliers were found in synthetic versus sensitive datasets, they | 200 GBP for research access, 1100 GBP for teaching access |

| | select the best synthetic batch to keep. | concluded that privacy was preserved. | |
| --- | --- | --- | --- |
| *IKNL Cancer dataset.* Synthetic dataset Netherlands Cancer Registry (NCR), Netherlands Comprehensive Cancer Organisation (IKNL). 2022. https://iknl.nl/en/ncr | | | |
| *Synthetic patients:* 60000<br><br>*Data overview*: single tumor sample, 46 variables<br><br>*Data types*: continuous, discrete and categorical | **Bayesian network:** A detailed explanation of the synthesis approach is not publicly available – according to direct correspondence, IKNL staff used PrivBayes to generate this dataset. | **Differential privacy: Privacy protections are assumed based on use of the PrivBayes tool with no further details publicly available.** | Application form for use in research, restricted sharing |

Our conclusion from this dataset review is that the huge opportunity for deploying synthetic health data derived from many digitalized health sectors in the EU is not being grasped. There also seems to be a general lack of application of the published synthetic datasets. Of the few datasets discussed above, only one seems to have been applied for other publications[49] and master's thesis work (the Synthea Covid-19 dataset). We strongly suspect this lack of activity is due to apprehensions over privacy preservation and the GDPR rather than lack of technical ability or interest.

**Deploying synthetic health data**

Synthetic health data offers a new, risk-reducing model for data access that particularly complements other growing access methods: national data biobanks and federated learning. Modern national biobanks containing data and biological samples are expanding in different forms in countries like the United States (All of US Research Program[50]), United Kingdom (UK Biobank[51,52]), and Finland (FinnGen[53]). These projects are devoting significant effort to multi-modal data de-identification, harmonization, and development of secure computing platforms - the last, unfortunately, rarely supports easy and flexible implementation of complex algorithms and pipelines. Federated learning is a developing technology that allows analysts to submit algorithms to spatially segregated local databases rather than a single root source of data and return analysis results without the researcher ever accessing the sensitive datasets themselves[54]. However, implementations are plagued by information loss and leaks, and also constrain the use of complex algorithms[54]. Synthetic health datasets can both contrast and augment the above models by supporting more flexible, less restrictive access to equally large, equally complex datasets. Additionally, synthetic data generation is theoretically highly tunable, offering different levels of privacy preservation for different application needs.

The main hurdle is that the regulatory environment is entirely unresolved around the dissemination of synthetic data in contrast to anonymized data and pseudonymized data[55]. We propose that, *in this early stage of synthetic data generation and deployment,* most synthetic health dataset access should be limited to approved, credentialed researchers in accordance with the majority of existing synthetic dataset deployment examples in the EU and UK. If deep learning methods have been used for generation from raw sensitive data without any further guarantees of anonymization by k-anonymity or differential privacy approaches, the datasets should be hosted within GDPR-compliant computing environments under guidelines that parallel those for pseudonymized sensitive health data. The critical advance that is immediately feasible to implement is the speed with which access to synthetic health data is granted. 'Exploratory research' should be an acceptable processing purpose within a minimal application for access to a synthetic dataset rather than constraining access solely to data justified by a specific preliminary hypothesis, and these applications should be fast-tracked for approval. Here, synthetic data can provide needed security and sensitive data access minimization for data authorities fearful of data leaks, and users will be restricted to researchers with prior experience using datasets at an appropriate sensitivity level.

A synthetic data solution is attractive because it immediately speeds up access to useful preliminary data for project design and algorithm prototyping. It simultaneously reduces the project evaluation queue for sensitive data access by weeding out unpromising research proposals at an early stage via synthetic data exploration results. The biggest advantage is that very little new infrastructure is needed for implementation outside of recruiting data science expertise in generating the synthetic datasets. These same data scientists could also assist in creating a facsimile of federated learning, where they transfer algorithms and pipelines developed on synthetic datasets by outside researchers to the sensitive data and return results to the researchers without needing to provide sensitive data access.

These pilot releases of synthetic health data to researchers will immediately expand our body of evidence for the potential uses and caveats of synthetic health datasets, speeding us along the path to appropriate deployment policies and dataset governance for an expanded user base. Determining which of these different conformations of synthetic data access and analysis requires testing (especially as number of users and datasets scales), but its immediate potential to enhance data-driven research should not be dismissed. We do not propose that synthetic datasets can fully replace the original sensitive data (certainly not for research publication purposes), but high-quality synthetic health datasets can uniquely support many useful applications before and alongside access to sensitive data. We see synthetic data as an important force multiplier – a way to increase access, challenge methods, and speed innovation in the health data science space.

**Conclusion**

Ultimately, the development of synthetic health data as a flexible tool to advance health data science should be managed under three guiding principles. First, the fidelity and utility of a synthetic dataset does not need to be perfect in order for the dataset to be useful. Sacrificing these qualities for others such as easy of evaluation and shareability due to low privacy risk is in some cases likely a worthy trade-off.

Second, synthetic datasets will likely always have limitations that make it difficult to replace real data in many important use cases. The generator and host of a synthetic dataset should clearly describe its intended use cases alongside expected limitations and caveats for downstream users. However, we fully expect the different potential use cases of synthetic data to continue to expand. And lastly, it is impossible to broadly guarantee that synthetic health data presents zero risk to privacy with respect to current and future technology. Instead, it is a question of mitigating risk to privacy under reasonable means of current and future attack.

Defined standards for risk assessment have not yet been implemented for synthetic data by higher data authorities. Data protection officers and legal advisors at data controlling authorities have limited time or capacity to engage with the mathematical underpinnings of the range of available privacy metrics for synthetic datasets. Meanwhile, responsible data scientists cannot in good conscience guarantee that there is zero privacy risk in the release of synthetic health datasets when modern deep learning methods are used to generate them. To alleviate these issues, there is a need for a set of consensus metrics that: 1) reflect concrete risk scenarios, 2) make realistic assumptions about the attackers background knowledge, 3) can be directly referred to when assessing GDPR compliance, and 4) are reliable for fully synthetic data with both continuous and discrete features. It is worth noting that converging on one single best practice may not be feasible or even recommended, as different evaluation metrics may be appropriate for different deployment scenarios, and so a consensus set of tailorable, interpretable, and coherent metrics should be designed with this in mind. The GDPR is not actually written inflexibly regarding patient privacy protection – data can be considered anonymized if identity and sensitive information cannot be recovered by 'reasonable measures.' Synthetic data practitioners, users, and data authorities must converge on a robust risk assessment framework that allows for different levels of risk according to proportionate gain derived from a given synthetic dataset and proportionate governance rules on who can access said dataset and under what conditions. This framework needs to be adaptable to different types of synthetic data alongside different use cases, and a consensus on the best quantitative metrics of privacy and fidelity should be reached based on supervised pilot projects that present systematic benchmarking results and comparisons for critique by all interested parties.

Currently, the EU is negotiating regulation regarding a European Health Data Space in order to provide uniform guidelines for collection, use, and sharing of sensitive health data[56,57]. We hope that this effort will demarcate a role for synthetic health data as a safe and reasonable mode of sharing information derived from sensitive health datasets within the EU. Meanwhile, we exhort practitioners to continue mapping the synthetic data solution space with reproducible studies using open access datasets so our understanding of this technology and its potential is transparently established and enhanced.

*Acknowledgments*
JAB and HL were funded by Erhvervsfyrtårn Life Science - Sund Vægt, a project supported by the EU Regional Fund (REACTRF-21-0024). HL was also supported by the Novo Nordisk Foundation grant NNF19SA0059129. SBV, JAB, AK, and MB were funded by the Novo Nordisk Foundation (NNF20OC0063268). AK was also supported by the Novo Nordisk Foundation grants NNF20OC0062606 and NNF20OC0059939 and the EU Horizon 2020 program (GenoMed4All project no. 101017549).